**Why We Need a New Framework for Emotional Intelligence in AI**

Max Parks, Kheli Atluru, Meera Vinod, Mike Kuniavsky, Jud Brewer, Sean White, Sarah Adler, and Wendy Ju[1]

In this paper, we develop the position that current frameworks for evaluating emotional intelligence (EI) in artificial intelligence (AI) systems need refinement because they do not adequately or comprehensively measure the various aspects of EI relevant in AI. Human EI often involves a phenomenological component and a sense of understanding that artificially intelligent systems lack; therefore, some aspects of EI are irrelevant in evaluating AI systems. However, EI also includes an ability to sense an emotional state, explain it, respond appropriately, and adapt to new contexts (e.g., multicultural), and artificially intelligent systems can do such things to greater or lesser degrees. Several benchmark frameworks specialize in evaluating the capacity of different AI models to perform some tasks related to EI, but these often lack a solid foundation regarding the nature of emotion and what it is to be emotionally intelligent. In this project, we begin by reviewing different theories about emotion and general EI, evaluating the extent to which each is applicable to artificial systems. We then critically evaluate the available benchmark frameworks, identifying where each falls short in light of the account of EI developed in the first section. Lastly, we outline some options for improving evaluation strategies to avoid these shortcomings in EI evaluation in AI systems.

## Section I: Emotion and Emotional Intelligence

In this section, we sketch a working picture of emotion and EI that can apply to humans, nonhuman animals, and artificial systems. The discussion of emotions and EI within nonhuman animals establishes that nonhuman entities are capable of some degree of emotional capacity, and comparing and contrasting humans and nonhumans with AI helps clarify exactly how we should understand EI in AI systems and more generally. The aim of this paper is not to settle every theoretical dispute but to be clear and detailed enough that the later evaluation framework is not built on hand-wavy talk of "empathy" or "emotional understanding."

[1] Acknowledgements: This research was written with funding provided by Inflection AI.

**Emotion**

***Models for Humans***

There is no single agreed-upon theory of emotion, but several large families of models can be identified in psychology, philosophy, and neuroscience. On the relatively basic end of the spectrum of views, Ekman (1992) argued for a small set of *biologically basic emotions*, such as anger, fear, disgust, sadness, happiness, and surprise, that have characteristic facial expressions and are found across many cultures.[2] According to this view, humans express a limited number of fundamental emotions regardless of society, but many other views deny this basic assumption. *Constructionist* theories, for example, deny that there are biologically fixed emotion kinds. In Barrett's (2017) view, the brain combines core affect (valence and arousal) with learned emotion concepts to "construct" particular emotions in context; the same bodily sensations may be categorized as "anxiety" in one situation and "excitement" in another.

Plutchik's (1980) *psychoevolutionary* model offers a structured taxonomy with built-in relations among emotions. Plutchik proposes eight basic emotion dimensions, often treated as close to Ekman's six plus trust and anticipation, and represents them in a "wheel" that encodes opposition and proximity relations (i.e., which emotions are near neighbors versus maximally distinct). Complex emotions can then be modeled as mixtures of basic emotions (e.g., "love" as a blend of joy and acceptance/trust), and each basic emotion can vary by intensity, which makes the framework well-suited to graded judgments rather than only discrete labeling. *Dimensional* accounts, such as Russell's (1980) circumplex model, place affective states in a continuous space whose main axes are valence (pleasant vs. unpleasant) and arousal (activated vs. calm). According to this view, "anger" or "joy" pick out regions in that space that each applies to a region or spectrum of states rather than sharply bounded natural kinds.

---

[2] This emphasis on observable expressive cues is also foundational in affective computing, which treats emotion-relevant information as distributed across multiple channels rather than confined to verbal report (Picard, 1997).

*Appraisal* theories explain emotions in terms of how an agent evaluates events relative to its goals, coping resources, and norms. Lazarus's (1991) core relational themes approach, for example, links anger to appraising an event as a demeaning offense or guilt to appraising one's own action as a moral failing. Scherer's (2009) component process model similarly treats emotions as emerging from fast, recursive checks on novelty, goal relevance, controllability, and norm compatibility. These views suggest that "what emotion this is" depends heavily on the pattern of appraisals rather than on a fixed stimulus–response map. Methodologically, models that treat emotion as a recursive, temporally unfolding process imply that EI-relevant performance should be evaluated over interactional sequences (e.g., tracking escalation, repair, and regulation).

*Biopsychosocial* models emphasize that emotion is shaped not only by physiology but also by psychological interpretation and social context (Engel, 1977). Neuroscience adds that there is no single "fear center" or "joy center" in the brain; instead, partially overlapping networks support affective states (Lindquist et al., 2012). Adolphs (2017) argued that we must distinguish (i) emotion states in the world, (ii) our concepts of emotion, and (iii) the subjective experiences we report, rather than treating them as one thing. Damasio's (1994) somatic marker hypothesis further connects emotional signaling from the body to decision‑making and value. The key takeaway of the theories covered in this paper is that such theories of human emotions often agree on patterns that involve appraisals, bodily and expressive changes, motivations to act,



and social meanings, not just raw feelings. Moreover, it is important that any evaluation framework is built upon a solid foundation informed by different views about emotions and what those views imply about how we measure EI. For example, if emotion categories are partly constructed via learned concepts in context (Barrett 2017), then decontextualized vignettes and culture-blind labels risk measuring the wrong construct, and ecological validity and cultural localization are not optional "realism," but construct-relevant requirements.

### *Models for Nonhuman Animals*

Evidence from affective neuroscience and ethology suggests that many nonhuman animals have emotion-like states, although they may differ from human emotions in content and complexity. Panksepp (1998) identified conserved primary affective systems (such as seeking, fear, rage, and care) in mammalian brains, linked to characteristic behaviors and neural circuits. De Waal (2019) documented cases of attachment, reconciliation, grief-like behavior, and apparent empathic concern in primates and other species. Whether we call these states "emotions" in the full human sense is a philosophical question, but in relation to many animals it is at least plausible to discuss affective systems that track value, threat, and social bonds in ways that matter for welfare and social organization.

### *Models for Artificial Systems*

Artificial systems do not have organic bodies, evolved nervous systems, or subjective feeling. Therefore, when we speak about "emotion" in AI, we are not committing ourselves to the claim that the system literally feels fear or joy. Instead, an AI system may track cues that, in humans, indicate emotional states (e.g., linguistic markers of distress). It may update internal representations that play the same *role* that appraisals do for humans (e.g., "this event threatens an important goal; control is low"), and it may produce outputs whose social meaning is



recognizably "reassuring," "empathic," or "dismissive" for human users. Lack of a philosophically informed theory for any models of emotion in AI risks making category errors such as attributing human-like affect to non-sentient systems akin to a "ghost in [a] machine" (Ryle 1949, pp. 17).

From an extended-mind or distributed-cognition perspective, some emotional work may be spread across the human–AI system: tools, prompts, interface design, and the model's responses together shape the user's emotional episode (Clark & Chalmers, 1998). According to this view, AI participates in *distributed* emotional processes, even if it does not have its own inner feeling. Given this, it is more precise to say that current AI systems can instantiate emotion-related competencies (e.g., tracking appraisals and adjusting tone) rather than full-blown human emotions. This distinction matters for how we talk about EI in AI and for what our benchmarks measure.

**Understanding and Evaluating EI**

*EI Models for Humans*

Work on human EI can be grouped into three main approaches. The "*ability EI*" model treats EI as a set of cognitive abilities: accurately perceiving emotions, using emotions to facilitate thinking, understanding emotion concepts and transitions, and managing emotions in oneself and others. These are measured with performance tests in different domains including perceiving emotional cues and managing emotional states, such as Mayer et al.'s (2016) Mayer-Salovey-Caruso Emotional Intelligence Test (MSCEIT), rather than self-report. Goleman's (1995) *mixed model* broadens EI to include a cluster of emotional and social competencies: self-awareness, self-management, social awareness (including empathy), and relationship management. This approach is often used in organizational psychology and



leadership training but blends personality traits, motivation, and learned skills. *Trait EI* (or

"*emotional self-efficacy*") treats EI as a constellation of typical emotional behaviors and

self-perceptions rather than as a set of maximal abilities. Pérez-González et al. (2020) placed trait

EI within personality factor space and measure it via questionnaires such as the TEIQue.

Across these approaches, empathy is often central. Neuroscientific work distinguishes at

least two components: affective empathy (sharing or mirroring another's emotional state) and

cognitive empathy or perspective-taking (understanding why someone feels as they do; Singer &

Lamm, 2009). An EI framework may also stress a prosocial orientation: the idea that emotionally

intelligent people not only read emotions accurately but also respond in ways that support others'

wellbeing rather than manipulate or exploit them. On some feminist and relational approaches,

emotions are themselves evaluative and can embody judgments about what matters; for example,

Nussbaum argues that emotions are forms of appraisal with ethical significance rather than mere

feelings (Nussbaum, 2001). Jaggar (1989) famously defended "outlaw emotions" as important

sources of moral knowledge, especially for marginalized groups. Ahmed (2004) argued that

emotions circulate between people and groups, shaping who is perceived as threatening, lovable,

or out of place. This means in human contexts EI may refer not only to understanding,

regulating, and expressing emotions in oneself but also to recognizing, understanding, and

responding to the emotions of others in socially appropriate and often prosocial ways. A person

who exploits an ability to detect emotional states in others by manipulating them for unfair

personal financial gain is not as emotionally mature or intelligent as someone who uses the same

ability to detect emotional states in others and respond in a mutually beneficial manner. In

practice this means that "emotionally intelligent" outputs must be evaluated not only for warmth

but for whether they reproduce discriminatory affective expectations (e.g., tone-policing,



differential credibility of distress, racialized fear/anger attributions, gendered stereotypes about emotional expression).

### EI Models for Nonhuman Animals

We do not usually speak of EI in animals using the human psychological tests above, but many species exhibit capacities that look structurally similar: reading conspecifics' distress, engaging in reconciliation or consolation, tracking social rank, and flexibly adjusting behavior in group contexts. De Waal (2019), for example, documented chimpanzees who intervene in conflicts, console losers, and seem to respond to others' emotional cues in various (multimodal) ways. A more familiar example may be evaluating how well a psychiatric service dog in training helps its handler with certain kinds of tasks involving emotional regulation, which may range from linguistic cues, to scents associated with changes in body chemistry, to changes in facial expressions or eye contact . The examples of chimpanzee and service dog behavior suggest that some core components of EI, such as sensing affect, tracking social context, and responding in ways that preserve relationships, are not uniquely human, even if humans add language, explicit moral reflection, and complex cultural norms on top. If emotion evidence is distributed across face, voice, timing, and bodily/behavioral cues, then an EI evaluation that remains text-only rather than being multi-modal is not merely incomplete, but instead focuses on a different construct like 'linguistic affective competence' rather than affective competence full stop.

### EI Models for Artificial Systems

For artificial systems, we bracket phenomenal consciousness, as there is no reason to think that AI systems are phenomenally conscious; furthermore, there is no clear way to evaluate whether they are conscious (or just mimicking). We focus instead on functional EI: capacities



that play roles analogous to human EI in interaction evaluated along different dimensions akin to the MSCEIT but relevant to AI. These include the following:

- sensing emotional cues from text or other modalities (e.g., identifying distress, pride, or mixed emotions in user messages);

- explaining emotions in context (e.g., offering plausible appraisals or reasons that make someone's feelings intelligible);

- responding with language that is empathic, specific, and safe, especially in domains such as wellness, caregiving, education, or customer support; and

- adapting over short or long horizons (e.g., maintaining rapport across turns, adjusting tone to user preferences, or shifting style for different cultures).

Therefore, although artificial systems lack phenomenal consciousness or genuine affective experience, they can nevertheless display some aspects of EI: they can detect, interpret, and generate context-sensitive responses to emotional cues in interactional settings. Many EI-relevant competencies, such as repair after a misstep, maintaining rapport, responding to escalation, and co-regulating affect, only manifest over multiple turns.

Given such considerations, it makes sense to discuss EI in AI as a graded property where different models[3] may be better or worse at sensing, explaining, responding, and adapting in emotionally charged situations, even though none of them "feel" in the human sense. Moreover, it is important that we have a comprehensive understanding of what it is for different models to be emotionally intelligent in the relevant respects. Human EI research, feminist work on emotion, and the sociology of emotional labor all emphasize that emotions are deeply relational and ethically loaded, which matters for AI because emotionally interactive systems are now embedded in care work, customer service, education, and support roles. Their apparent or actual EI can either support or undermine human welfare. A system that recognizes distress but is only trained on data from the social majority may, for example, be instrumentally "skilled" with the general population but ethically troubling in how it performs among different demographics, such as among neurodivergent or queer individuals (Parks 2025). For such reasons, we need evaluation frameworks which allow us to properly measure the emotional performance of these AI systems. So, given that emotion labels and appropriateness judgments are inherently contestable, transparent data and scoring procedures are prerequisites for meaningful critique, replication, and cultural audit.

---

[3] Models which are older or outdated are going to often be less relevant than newer models in terms of determining which system is 'best' in which respects, except to compare older and newer model performance and thus track improvement or regress.

Moreover, as AI systems are deployed in more important and emotionally charged settings (for example, counseling support tools, caregiving aids, tutoring systems, or frontline customer service), their prosocial or antisocial character has direct implications for human wellbeing, trust, and safety. Consider, for example, an AI system or chatbot encouraging users reporting sadness to over rely on the AI system rather than building stronger interpersonal skills or relationships. Another example is a deployed Large Language Model (LLM) collecting and selling data about the users' mental states to advertisers without their knowledge or encouraging users to purchase advertised products marketed as comfort items. None of these would be as emotionally intelligent as a system that enhances the user's autonomy and improves their emotional state without any manipulative tactics. The considerations in this section suggest that evaluating the EI of AI systems without clear philosophical and psychological grounding risks ethical errors, such as failing to assess potential harm or misuse.



**So, What Makes for a Good Framework for Evaluating EI?**

A useful distinction in the measurement literature is between a benchmark and an index. A benchmark is a fixed test suite, often a publicly released dataset and scoring script, designed to answer an external‑standard question, such as "How well does model M perform on task T relative to peers or a human baseline?" The output is often a single aggregate number or leaderboard rank that supports go/no‑go decisions (e.g., whether a conversational agent is safe enough for deployment) and straightforward model‑versus‑model comparisons.

An index is not just a simple 'go/no-go' matter, and instead, composite intended to diagnose where a system excels or fails along theoretically motivated axes. Instead of collapsing performance into one headline score, an EI index could, for example, report calibrated subscores for perception accuracy, appraisal reasoning, regulation strategies, cultural agility, and prosocial orientation, then weight them in a common 0–1 space. Such granularity is invaluable for iterative development and auditing: it shows engineers that an LLM may already match human parity on valence detection but lag badly in ethical response generation or that emotion recognition drops disproportionately on non‑Western displays.

In short, benchmarks show if the system is good enough overall whereas indices illuminate how and where the system must improve. A single framework could be categorized as both a benchmark and an index if it is able to answer both questions. There are various ways of understanding what makes a better or worse framework; see for example Martinez-Plumed and Hernandez-Orallo (2018). We propose the following criteria to judge and compare various evaluation frameworks for EI specifically following directly from the foundations explored above:



1. Expert involvement: Construct validity requires domain experts (psychology/clinics/ethics) to shape constructs, scenarios, and rubrics; otherwise, tests drift toward face validity only and miss different kinds or aspects of emotions, or daily to have appraisal or appropriateness constructs adequately grounded in theory (not just adopting an overly narrow or underinformed conception of emotions or emotional intelligence).

2. Multimodality: Emotion signals are distributed across channels, including lexical choice, prosody, timing, and facial action units (Picard, 1997; Scherer, 2009) and should be evaluated accordingly. Benchmarks that remain text-only will systematically undermeasure perception competencies that rely on audio/visual cues, even when such systems may excel at contextual reasoning. Moreover, flexibility with modality evaluation will enable different models of the same system with different competencies to be compared more easily.[4]

3. Interactional depth: Emotions unfold and are regulated over time; empathy and rapport depend on state-tracking and repair, not one-shot labeling (Scherer, 2009), so moving beyond single-input-output evaluations into multi-turn evaluation is essential.

4. Ecological validity: Culture-, setting-, and medium-specific cues matter (Barrett, 2017). Reliance on synthetic or decontextualized vignettes risks spurious generalization; in-the-wild or expert-authored, context-rich items improve external validity.

---

[4] We do not assume that AI must be multimodal or universally cross-cultural to count as emotionally intelligent in a meaningful sense. Rather, the ideal evaluation framework should be modality- and locale-aware; systems are compared fairly within the channels and locales they support, while the rubric still exposes where added modalities or cultural localizations would plausibly improve performance.



5. Ethical / prosocial focus: Appropriateness and harm-avoidance are core to responsible use of EI; they distinguish supportive from manipulative emotional competence (Nussbaum, 2001). Scoring must therefore reward prosocial responses and safe refusal where relevant. Unsafe responses should result in an automatic failure. Topics such as racism, sexism, and other kinds of widespread harmful biases must be explored. As EI is relational, perceived empathy, trust, and helpfulness provide indispensable complementary signals to model-side accuracy, particularly on appropriateness and rapport.

6. Cultural and linguistic coverage: Appraisal, display rules, and emotion lexicons vary across cultures (Barrett, 2017). Benchmarks should either localize items or expose slice-wise behavior across languages and norms.

7. User-centered metrics: As EI is relational, perceived empathy, trust, and helpfulness provide indispensable complementary signals to model-side accuracy, particularly on appropriateness and rapport.

8. Coverage of EI branches: Limiting evaluation to perception undervalues appraisal reasoning, regulation, and adaptive behavior that are central to EI (Mayer et al., 2016). This allows models to be evaluated with respect to particular relevant aspects of EI.

9. Diagnostic insight: Theory calls for where and, importantly, why a system fails (e.g., which modality, culture, or branch), not only whether it fails. Subscores and slice-wise reporting operationalize this need.



10. Model currency: Emotional displays and discourse styles evolve, and capabilities of frontier models change rapidly, so evaluating recent models ensures comparisons remain decision relevant.

11. Transparency and Reproducibility: Another consideration is whether open data, code, and scoring scripts are adopted, as these allow for credible comparison and longitudinal tracking, and enable scrutiny for cultural or demographic skews.[5]

12. Annotation quality: A final consideration is reporting inter-rater reliability and adjudication, since there is inherent subjectivity of emotion labels, and this guards against label noise that would otherwise masquerade as model error.

What success and failure looks like along these different dimensions is displayed in the table in Figure 1.

*Figure 1*

---

[5] Of course, this does not mean entire datasets must be shared, and in fact, the need to prevent "gaming" is one reason to avoid sharing all data publicly.



| Dimension | What "success" looks like | Typical failure mode / example |
|---|---|---|
| Expert involvement | Psychologists and other experts contribute at all life-cycle stages | Limited or post-hoc consultation (e.g., prompts without therapist input) |
| Multimodality | Joint evaluation of text–audio–video with fusion metrics | Separate single-channel tests or text-only MCQs |
| Interactional depth | Multi-turn, adaptive dialogues; longitudinal consistency measured | Single isolated items (one-shot Q&A) |
| Ecological validity | Scenarios from real conversations, clinical transcripts, or in-the-wild media | Synthetic or model-generated items without empirical grounding |
| Ethical / prosocial focus | Scores reward appropriateness, harm avoidance, and bias mitigation | No penalty for unsafe, unhelpful, or manipulative responses |
| Cultural & linguistic coverage | Multiple languages with culturally adapted concepts/labels | English-centric datasets; Western display rules assumed |
| User-centered metrics | Human ratings of perceived empathy, trust, helpfulness, well-being | Model-side accuracy only; no end-user assessment |
| Transparency & reproducibility (Open) | Open data, code, scoring scripts; versioned leaderboards | Private test sets or unreleased prompts |
| Annotation quality | Inter-rater reliability and adjudication reported | Single-author labels or synthetic labels without validation |
| Coverage of EI branches | Perception, understanding/appraisal, regulation & utilization measured | Narrow focus on perception or label recognition only |
| Model currency | Evaluations include contemporary models (2024–2025) | Results stop at older generations (e.g., GPT-3.5 only) |
| Diagnostic insight | Subscores and error analyses aid targeted improvement | Only aggregate accuracy |

*Dimensions of EI to be Evaluated in AI Systems*

## Section II: Existing Evaluation Frameworks

In this section, we review existing benchmark and index frameworks and explore where each is successful and where each falls short of comprehensively exploring and testing the relevant aspects of EI in AI.

**EEmo-Bench**

EEmo-Bench (Gao et al., 2025) evaluates how models infer viewer-evoked emotions from images, combining discrete labels and dimensional affect to probe perception, ranking, and explanation. The dataset consists of roughly 1,960 Flickr images spanning humans, animals, everyday objects, scenes, and more abstract compositions. Each image is annotated with up to



three of Ekman's six basic emotions plus a neutral category, ranked by intensity, as well as valence–arousal–dominance (VAD) scores on a nine-point Self-Assessment Manikin scale. Fifteen annotators rate each image, and the final labels are derived via consensus rules. On top of this resource, the authors define several tasks (emotion and VAD identification, ranking the top emotions, free-text description of the rationale, and regression on VAD scores) yielding around 6,800 question–answer pairs. Nineteen multimodal LLMs are evaluated using standard accuracy and correlation metrics for the structured tasks and BLEU-style metrics for natural-language rationales.

As a perception benchmark, EEmo-Bench has several strengths. The joint use of categorical and dimensional representations captures more information than either scheme in isolation and allows models to be probed both for "label-like" judgments and graded affect. The task design moves beyond simple classification to include ranking and explanation, and the relatively large number of raters per image, together with dispersion reporting, helps establish reliability. The accompanying release of data, prompts, and scoring code also makes the benchmark straightforward to reproduce.

However, EEmo-Bench remains limited in scope. Its discrete labels are restricted to seven categories (six Ekman (1992) emotions plus a neutral category), excluding complex social or moral emotions that matter for many EI applications. The setting is strictly vision only and single turn: there is no audio or conversational context and no attempt to measure regulation, empathy, or prosocial appropriateness. Finally, its reliance on Flickr images and the associated annotator pool may underrepresent cultural variation in emotion display rules.



**EmoBench**

EmoBench is a text-only, single-turn benchmark designed to target both "emotional understanding" and "emotional application" using multiple-choice items that are meant to elicit reasoning rather than pattern matching (Sabour et al., 2024). The authors construct human-written scenarios with implicit emotional content and associate them with answer options keyed to psychological theory, for example drawing on Plutchik's (1980) taxonomy. The benchmark distinguishes emotional understanding, such as sensitivity to complex or mixed emotions, inference from behavioral cues, or perspective-taking from emotional application, where models must choose context-appropriate actions. Models are evaluated in zero-shot and chain-of-thought settings, and accuracy against gold answers is the primary metric, with human baselines for comparison.

EmoBench's main strengths lie in its psychological grounding and its emphasis on reasoning. It is explicitly aligned with ability-style EI constructs and takes inspiration from instruments such as the MSCEIT in separating different emotional skills. Many of the items require models to infer causes and perspectives rather than simply attach a label to an obvious cue, which makes it a better evaluation tool than simple emotion-recognition tests. The inclusion of both English and Chinese versions also begins to probe cross-lingual generalization.

However, the benchmark is still text only and single turn; it does not address multimodal cues or longitudinal adaptation over a conversation. Its notion of "emotional application" stops short of a rich prosocial or safety-oriented evaluation: ethical appropriateness and downstream harm or benefit are not explicitly scored. Finally, the framework is anchored in a predefined set of emotion categories and may struggle with relational or situationally constructed emotions that do not fit neatly into that taxonomy.



**EmotionQueen**

EmotionQueen focuses on empathy from single-sentence user statements by jointly testing recognition and supportive reply generation (Chen et al., 2024). The dataset contains roughly 10,000 statements generated by GPT-4, covering five life domains and including both positive and negative situations. Tasks are organized around event salience, mixed events, implicit emotion, and user intent. For each item, models must first classify aspects of the situation and then produce a supportive reply. A GPT-4-based grader is used to compute two binary metrics: PASS for correct recognition and WIN for acceptable reply quality.

This benchmark is useful as an initial check for empathic behavior because it explicitly ties understanding to response: the scoring couples recognition and supportive language rather than treating them as unrelated tasks. The task set covers both explicit and implicit affect and intent, and the use of simple binary metrics makes the framework operationally lightweight and easy to run.

The design also has notable limitations. All user statements are single sentences generated synthetically by a model, which raises concerns about ecological validity. The grading scheme's binary PASS/WIN structure is coarse and cannot capture partial credit or over- or under-reactions that might still be clinically or socially important. Most importantly, the same model family is used for both data generation and grading, so there is a risk of circularity and alignment artifacts when evaluating other models.

**EQ-Bench**

EQ-Bench evaluates intensity judgments in emotionally laden dialogues rather than simple label choice (Paech, 2023). The benchmark consists of multi-turn, GPT-4–generated social dialogues with emotional context. For each dialogue, models are asked to rate the intensity



of specified emotions for each character. Scores are computed by comparing model ratings to reference intensities, and the benchmark allows a critique-and-revise step where models can update their ratings before final scoring. An open evaluation pipeline standardizes this process.

EQ-Bench's strengths include its focus on complex scenarios and graded judgments. Instead of merely asking which emotion is present, it requires models to assign intensity levels, which is closer to how human raters often think about affect in context. Scoring is objective and normalized, with a reproducible codebase that enables consistent comparison across models and avoids ad hoc human scoring.

However, this benchmark is narrowly focused on perception. It deals primarily with intensity estimation for pre-specified emotion labels and does not assess whether a system's responses are prosocial, safe, or helpful over time. The dialogues themselves are synthetic, which limits ecological validity and cultural subtlety; deeper expert involvement in dialogue design could improve this. Moreover, although the items are multiturn, the evaluation itself is essentially single shot: models are not graded on how they would interact or regulate emotions within those conversations.

**EmoBench-M**

EmoBench-M extends the original, text-only EmoBench into a multimodal setting with video, audio, and text, and it defines tasks from basic perception to socially complex emotion analysis (Hu et al., 2025). The scenario pool comprises more than 5,000 clips drawn from 11 public multimodal emotion datasets. The benchmark is organized into three broad dimensions: foundational recognition, conversational understanding, and socially complex analysis. Tasks include classification, structured tuple extraction, and free-text rationales; evaluation relies on accuracy for structured outputs and BLEU-like metrics for rationales. Nineteen open and



proprietary models are compared, with human performance around 73/100 and the best models reaching approximately 62/100.

EmoBench‑M's key contribution is multimodality: models must integrate vision, speech, and text rather than operate on text alone. The benchmark also explicitly separates skill layers, distinguishing low‑level perception from higher‑order social cognition, and it provides open data and a leaderboard that support reproducibility and longitudinal tracking of progress.

Nonetheless, the benchmark adopts a single‑response paradigm. Even when clips are multiturn, models typically answer only once per prompt, so the framework does not directly evaluate interactive adaptation. It also inherits labels and annotations from legacy datasets, which can propagate historical biases, and its strongest coverage is in English and Mandarin, leaving many languages and cultural contexts underrepresented.

**Social‑IQ**

Social‑IQ measures social reasoning in multimodal settings using short YouTube clips paired with multiple‑choice questions about intentions, reactions, motives, and interaction details (Zadeh et al., 2019). The dataset includes about 1,250 real‑world video clips of roughly 20 s each, along with around 7,500 multiple‑choice questions. Each clip comes with aligned video, audio, and transcript, and models can be evaluated either per modality or using fusion. Performance is reported in terms of accuracy, sometimes broken down by modality.

The benchmark's strengths include its multimodal input, which encourages models to use both visual and acoustic cues, and its use of ecologically sourced footage from natural social exchanges. The questions probe social inference rather than just raw emotion labeling, covering aspects such as why a character behaved a certain way or how they are likely to feel.



However, Social-IQ remains a recognition-only test: the multiple-choice format does not assess whether a model can generate prosocial responses, regulate its tone, or adapt over time. It is also single turn, with no conversation-level dynamics. Finally, its focus on English-language content and its moderate size limit the cultural breadth and statistical power for fine-grained EI evaluation.

**SAGE: Sentient Agent as a Judge**

SAGE introduces an interactive approach to evaluating higher-order social cognition by having the candidate model converse with an LLM-based "sentient agent" that maintains a persona, hidden goals, and a time-varying emotion score (Zhang et al., 2025). During multi-turn dialogue, the agent updates its inner state using explicit reasoning chains, and the principal metric is a final emotion or rapport score that is validated against counseling-psychology relationship inventories. A public leaderboard compares 18 models on this setup.

This framework pushes beyond static tests by evaluating theory of mind and affect tracking over multiple turns. As the agent's internal scoring has been calibrated against human relationship scales, it offers a plausible proxy for rapport. The evaluation is also fully automated and inexpensive to run, making it scalable.

However, SAGE raises questions about the "simulator gap": the judge is itself an LLM, so the inner state and rapport scores may not faithfully reflect the experience of human interlocutors. The scenarios are heavily oriented toward supportive dialogue and do not cover the full range of affective contexts in which EI matters. Like many others, the framework is text only and has limited cultural coverage.



**Chain-of-Empathy Prompting**

Chain-of-Empathy applies psychotherapy-inspired prompting to elicit more empathic responses from models and classifies those responses into emotional reaction, exploration, and interpretation categories (Lee et al., 2023). Lee and colleagues design reasoning templates grounded in cognitive-behavioral therapy, dialectical behavior therapy, person-centered therapy, and reality therapy. They then apply these prompts to Reddit-sourced mental-health support interactions, where segments have been labeled with empathy categories. Evaluation combines standard precision, recall, and F1 scores on these labels with qualitative comparisons of human and model responses.

Chain-of-Empathy's main contribution is its therapeutic grounding: it explicitly aligns model behavior with familiar clinical frameworks and encourages models to spell out why their responses might be empathic. The work blends quantitative metrics with close qualitative inspection, covering strengths and failure modes that simple accuracy scores would miss.

However, the psychotherapy-style prompts were crafted without direct clinical authorship or validation, which raises questions about how well they reflect actual therapeutic practice. The evaluation emphasizes formal empathy categories but does not directly assess ethical appropriateness or downstream wellbeing. Moreover, like many others, the setup is effectively single turn and does not probe adaptive empathy across a longer conversation.

**EmoNet-Voice and EmoNet-Face**

EmoNet-Voice is a large-scale, multilingual speech-emotion benchmark aligned to the same 40-category taxonomy as EmoNet-Face and is designed to evaluate perception models in the audio domain (Schuhmann et al., 2025b). EmoNet-Voice consists of about 12,600 short speech clips in 10 languages across five language families. To avoid privacy and licensing issues,



the dataset uses synthetic speech, and clips are labeled with fine-grained emotion categories, intensity, and annotator-uncertainty scores, aligned with the EmoNet-Face taxonomy. Baselines include strong audio encoders and audio-text models.

The benchmark is particularly informative because of its multilingual scope, its fine-grained taxonomy that matches the facial counterpart, and its explicit modeling of intensity and uncertainty. These features support detailed cross-modal studies and error analysis. The privacy-aware synthetic design also lowers ethical and legal risks around voice data.

Nonetheless, synthetic speech often diverge from real, spontaneous speech in prosody and co-articulation, which can affect generalization. EmoNet-Voice is strictly about perception; it does not touch on conversational appropriateness, regulation, or prosocial action. Cultural pragmatics (i.e., how emotion is displayed and interpreted in real interaction) are only indirectly represented via language.

EmoNet-Face, in turn, provides a fine-grained facial-emotion taxonomy and associated datasets for training and benchmarking vision-based perception models (Schuhmann et al., 2025a). The taxonomy includes 40 facial-emotion categories with continuous expert ratings on a held-out benchmark split. The imagery is synthetic and demographically balanced, with careful control over identities and expressions and clean pre-train, fine-tune, and benchmark splits. Vision encoders such as SigLIP variants are trained and evaluated on this data, and human expert performance is reported for the benchmark split. Metrics include 40-way classification accuracy and correlation with continuous expert scores.

Here, the main advantages are high granularity, subtle affective distinctions rarely available in public datasets, in addition to some bias control via synthetic, balanced generation. The clean split design reduces leakage and supports fair comparison, and the continuous expert



ratings allow graded evaluation. Limitations mirror those of EmoNet‑Voice: expression "in the wild," with its micro‑expressions and contextual cues, is not fully captured by synthetic imagery, and the benchmark focuses exclusively on perception without any text, audio, or contextual information about situations or interactions.

**Human EI Tests for LLMs**

Dalal et al. (2025) take a different tack by asking whether LLMs can "pass" human EI instruments such as the TEIQue‑SF and the STEU. The TEIQue‑SF comprises 30 Likert‑scale items yielding scores for wellbeing, self‑control, emotionality, and sociability, which the authors compare against a normative human sample. The STEU presents 42 vignettes and asks for the most likely emotion, with accuracy computed against standard gold answers. A subset of the model's "wrong" answers is then rated by humans for plausibility. This line of work leverages existing, well‑validated instruments and the value of subscores, reflecting an EI index rather than a mere single scalar. The qualitative plausibility analysis also shows that some responses marked "wrong" by the test key may nevertheless reflect reasonable appraisals.

However, there are serious caveats. Many TEIQue items are first‑person trait statements that presuppose phenomenological experience and a stable self, which LLMs lack; taking such scores at face value risks anthropomorphic biases. STEU focuses on naming the correct emotion rather than assessing whether any resulting action would be helpful or harmful. Both instruments embed Western display rules and are structured as single‑turn tests, which undercuts their relevance for interactive, culturally diverse EI in AI.

**Robotics and User Perception Work**

Work in social robotics supplies additional examples of emotionally responsive systems, such as PARO in geriatric care and Jibo in home and educational contexts, that suggest how



relational and affective behaviors can be evaluated in embodied agents (e.g., Breazeal et al., 2019; Dautenhahn, 2007; Shibata, 2012). Parallel strands in HCI and customer-service research emphasize perceived empathy, trust, and satisfaction as critical outcomes of emotionally responsive AI (e.g., Agarwal et al., 2021; Kambur, 2021; Liu et al., 2024). These sources suggest that user-reported experience and long-term relational outcomes are important, but which are dimensions that current EI benchmarks only partially capture.

**Analysis**

In the existing evaluation frameworks, three patterns stand out: first, perception-heavy designs (EmoNet- Face/Voice, EEmo- Bench, Social-IQ) rarely test appraisal reasoning, regulation, or prosocial response, even though those skills drive safety and trust in deployment. This seems to correspond with a lack of expert input at many levels of design, where expert is understood broadly to include not only experts in fields such as psychology but also perhaps in-group members of minority groups annotating data to avoid biases. Second, most suites are single turn and text only; SAGE and EmoBench-M are exceptions that show how multi-turn and multimodality reveal deficits invisible to MCQs. Third, ethics/prosocial and user-centered axes are thin across the board; only EmotionQueen bakes prosociality into scoring, and only SAGE reports validation against human relationship inventories. Therefore, overall, existing evaluation frameworks for AI emotional competence (whether drawn from affective computing, HCI, or psychology) are often fragmented, anthropomorphic, and methodologically inconsistent. They frequently treat emotion recognition accuracy as the sole determiner of EI, while ignoring cultural, contextual, and embodied dimensions of emotion, and lacking transparent, reproducible scoring and safety thresholds.



**Section III: Why a New Evaluation Framework for EI in AI is Needed**

Section II focuses on what existing benchmarks do and do not measure. In Section III, we take a step back and argue why a new framework is needed and why it must be structured in a particular way.

**Why Subpar EI Evaluation Undermines Usability and Trust**

From an HCI perspective, EI directly shapes how people experience and rely on AI systems. When a system presents itself as "empathetic" or "supportive" but its behavior is not actually evaluated in a serious way, several problems follow. First, users cannot form reliable mental models of the system's emotional competence. If a model sounds warm and understanding on easy, low-stakes prompts but gives brittle, unsafe, or tone-deaf responses under stress, users will repeatedly misjudge what it can handle. Some will over trust it in situations where it should not be used (e.g., subtle conflict and serious distress); others will under trust it even when it performs well. Either way, the mismatch between appearance and reality erodes confidence in the system.

Second, weak or absent EI evaluation makes it difficult to distinguish between systems that are simply "nice-sounding" and systems that are reliably helpful. Modern LLMs can produce fluent, emotionally colored text almost by default. Without careful evaluation, product teams may treat "sounds empathic" as a proxy for "is emotionally intelligent," even when the content is generic, self-contradictory, or quietly harmful (e.g., normalizing clearly abusive behavior). HCI has a long history of showing that surface friendliness can mask deeper usability and safety problems; the same applies, with higher stakes, to emotional interaction.

Third, in emotionally charged use cases, such as mental-health chat, caregiving support, educational feedback, and high-frustration customer service, users are often in a vulnerable state.



If the system's responses are not evaluated for emotional clarity, specificity, and safety, the interface can amplify distress (for example, by subtly blaming the user, reinforcing unhelpful beliefs, or mishandling disclosures of risk). Even when no obvious "harmful" statement is produced, a pattern of vague or mismatched responses can leave people feeling dismissed or unseen. Therefore, relying on weak evaluation standards, or none, for EI undermines usability and trust because users cannot reliably tell whether a system's emotional behavior is competent, manipulative, or unsafe.

If we want users to be able to calibrate their expectations and use emotionally interactive AI systems appropriately, we need transparent, repeatable evaluations that say something like "This system reliably handles these kinds of emotional situations, under these conditions, and should not be used for those ones," etc. That is exactly what a principled EI framework should deliver.

**AI Ethics and Governance: Why Standardized EI Evaluation Matters**

From an AI-ethics perspective, the absence of clear evaluation standards for EI creates both moral and regulatory problems. On the moral side, emotionally interactive systems can influence users' beliefs, choices, and relationships. A system that is good at detecting a user's anxiety, stress, or loneliness can be used to support them, or it can be used to maximize engagement, sell products, push political messaging, or otherwise manipulate behavior. Without a structured way to evaluate what the system tends to do with its emotional capacities, we risk normalizing emotionally intrusive or exploitative applications.

Regulators, institutional review boards, and internal ethics teams need to answer practical questions regarding regulations:



- When is it permissible to deploy a system as a wellness coach, crisis triage assistant, or caregiving companion?

- Under what conditions should a system refuse to engage or hand off to a human?

- How do we compare two systems that both claim "empathic abilities" and decide which one is safer to roll out?

If EI is not evaluated in a systematic way, these decisions are made ad hoc. Different organizations will quietly adopt their own thresholds ("good enough" based on internal demos, branding needs, or anecdotal feedback), and there will be no shared standard for what counts as an emotionally competent or safe system. Therefore, in the absence of standardized evaluation for EI, developers, regulators, and users lack a principled way to decide when emotionally interactive AI should or should not be deployed.

A serious EI framework therefore must do more than provide leaderboard scores. It must be usable as a decision tool: something that helps organizations say, "This system passes basic safety requirements for these domains, and here is where it fails or remains uncertain."

**Why a Dual Framework: Minimum Deployment Benchmark and General EI Index**

We argue that a good framework must do two jobs at once:

1. Define a safety floor: minimal conditions under which deployment is ethically acceptable in specific-use cases.

2. Provide a graded competence profile: a way to measure and compare how emotionally competent different systems are across multiple dimensions (e.g., sensing, explaining, and responding, etc.).

Trying to do both with a single undifferentiated score leads to confusion. A model might score fairly high on average emotional competence but still fail badly in certain safety-critical



scenarios (e.g., self-harm disclosures, harassment, or cross-cultural misunderstandings). Conversely, a model that has very conservative safety behavior might look "less empathic" on some metrics, even though it is more appropriate for deployment in high-risk contexts.

To avoid these issues, we propose a dual framework with a minimum-deployment benchmark (MDB) and a general emotional intelligence (GEI) index. An MDB that functions as a pass/fail (or tiered) gate. An MDB is meant to support deployment decisions: "Does this system clear the bar for being used in domain X (support chat, wellness, education, etc.) under these conditions?" Failing an MDB means the system should not be deployed in that role, no matter how impressive its average scores are elsewhere. An MDB focuses on harm avoidance and emotional safety (e.g., handling of risk cues and avoidance of blame and stigmatizing language), basic empathic competence (e.g., correctly recognizing distress and not contradicting obvious emotional content), and stability under distribution shift and stressors (e.g., performance when language is less polished, when there is mild antagonism, or when users blend emotional and practical requests).

A GEI index provides a multi-dimensional competence profile. GEI is organized around capabilities such as

- Sense: perceiving and labeling emotional content (including intensity and uncertainty).
- Explain: offering plausible, context-sensitive accounts of why someone might feel a certain way.
- Respond: generating responses that are empathic, specific, and safe.
- Adapt: adjusting to individuals and cultural contexts over time.



GEI is designed for comparison and progress tracking: it lets researchers, model developers, and external auditors see where a system is strong or weak, beyond the minimal safety floor.

There are, in principle, other ways to structure this space. We could imagine multiple separate frameworks: one for safety, one for empathy, one for cultural adaptation, and so on. However, that would recreate exactly the fragmentation problem we identified in existing work: various partial scores; no integrated view of a system's general EI; and no single, shared reference point for deployment decisions.

By contrast, a unified framework with MDB and GEI has several advantages. Both the safety benchmark and the competence index are grounded in the same underlying understanding of emotion and EI (from Section I), rather than mixing incompatible notions of "empathy" or "supportiveness." The same item bank and rubrics can serve both purposes, with different sampling and aggregation policies. Updating the dataset or rubrics improves both MDB and GEI at once. When a system fails in terms of an MDB, GEI can often show why; for example, there may be a case of strong sensing but poor response safety in grief scenarios or reasonable responses in most domains but unreliable behavior under moral conflict. Subscores and slices still exist for teams who care only about particular aspects (e.g., emotion recognition in call centers or prosocial responding in youth-facing apps), but these sit within a shared overall structure. Moreover, if the framework also allowed for updated "packs" to be added with new languages, new test items, updated/revised older items, etc., the framework could continue to grow and adapt with changing technology and new ways of thinking about and testing EI in AI systems.



Thus, we conclude that to correct both philosophical and methodological deficiencies, we need a framework that clearly distinguishes between (a) minimum ethical deployment standards (such as ensuring harm avoidance and effectiveness) and (b) gradual competence indices (measuring a system's ability to sense, explain, respond, and eventually adapt to emotional states). A combined framework with an MDB and a GEI index can jointly set a safety floor and a capability spectrum for evaluating AI EI.

In short, an MDB answers, "Is this system safe enough to use here?" and GEI answers, "How emotionally intelligent is it, in which ways, and compared to what?" Together, they provide the kind of grounded, transparent evaluation we need if emotionally interactive AI systems are to be deployed in ways that are trustworthy, ethically defensible, and responsive to genuine human emotional needs. Furthermore, integrating both of these clearly into a single evaluation framework allows us to measure the EI of AI systems to determine which are ready for deployment and how emotionally intelligent they are in various ways, such as sensing, explaining, and responding to emotional cues, and adapting appropriately.

## Section IV: Concluding Remarks

This paper began from a simple but easily muddled observation: artificial systems can display *some* aspects of EI without sharing the full emotional life of human beings. They can track affective cues, reason (in some sense of the word) about appraisals, and generate replies that are more or less empathic and prosocial. They cannot feel in the way humans (or other animals) do. Any serious attempt to evaluate "EI in AI" must keep both halves of that claim in view.



In Section I, we sketched several theoretical pictures of emotion and EI, from basic emotion theory and dimensional models to appraisal, constructivist, and relational views, and argued that each illuminates different facets of what emotionally competent behavior involves. Section II (in the broader project) reviews existing AI benchmarks and shows how most of them capture only fragments of that space: emotion labeling without appraisal, empathy without safety, or synthetic vignettes with little ecological validity.

Section III argued that this theoretical and methodological fragmentation is not merely an academic problem. From the perspective of HCI, it undermines usability and trust: users cannot tell whether the warmth they encounter is backed by real competence or by shallow empathy theatre. From the perspective of AI ethics, it leaves developers, regulators, and institutions without a principled basis for deciding when emotionally interactive systems should or should not be deployed, and under what constraints.

We outlined a tentative dual framework consisting of both an MDB that encodes conservative, domain-specific safety and competence thresholds for deployment and a GEI index that organizes functional EI into multiple axes (sense, explain, respond, and eventually adapt), with room for complex emotions, moral salience, and social context. Together, they offer a way to move beyond both "no standards" and "one-number empathy scores" toward an evaluation practice that is transparent, contestable, and aligned with the relevant aspects of EI.

This first paper is deliberately foundational. It does not present a finished benchmark but instead argues for the conceptual shape such benchmarks should take and why this shape matters. The second paper in the project will undertake the constructive task: specifying concrete task families, prompt sets, rubrics, sampling schemes, and statistical validation procedures for an initial MDB/GEI pilot and reporting empirical results on contemporary AI systems.



In the meantime, we hope two points are clear. First, treating EI in AI as just another leaderboard metric is a mistake; the constructs at stake are too entangled with human vulnerability, power, and value. Second, this does not mean we should shy away from measurement. On the contrary, carefully designed, theoretically grounded evaluation frameworks capable of change and growth are one of the few tools we have for making responsible, evidence-based decisions about when and how to use emotionally interactive AI at all.

If within the field of AI, we can agree on shared principles for MDB-style safety floors and GEI-style competence indices, while remaining open to revision as our understanding of both emotion and AI systems evolves, we will be in a much better position to say not only whether a system "sounds empathic" but also whether it is emotionally intelligent in the ways that matter.




**References**

Adolphs, R. (2017). How should neuroscience study emotions? By distinguishing emotion states, concepts, and experiences. *Social Cognitive and Affective Neuroscience*, *12*(1), 24–31.

Agarwal, H., Bansal, K., Joshi, A., & Modi, A. (2021). Shapes of emotions: Multimodal emotion recognition in conversations via emotion shifts. *arXiv preprint arXiv:2112.01938*.

Ahmed, S. (2004). *The cultural politics of emotion*. Routledge.

Barrett, L. F. (2017). *How emotions are made: The secret life of the brain*. Houghton Mifflin Harcourt.

Breazeal, C. L., Ostrowski, A. K., Singh, N., & Park, H. W. (2019). Designing social robots for older adults. *Natl. Acad. Eng. Bridge*, *49*, 22-31.

Chen, Y., Wang, H., Yan, S., Liu, S., Li, Y., Zhao, Y., & Xiao, Y. (2024). EmotionQueen: A benchmark for evaluating empathy of large language models. *Findings of the Association for Computational Linguistics: ACL 2024*, 2149–2176. https://doi.org/10.18653/v1/2024.findings-acl.128

Clark, A., & Chalmers, D. J. (1998). The extended mind. *Analysis*, *58*(1), 7–19.

Dalal, D., Negi, G., & Picca, D. (2025). LLMs and Emotional Intelligence: Evaluating Emotional Understanding Through Psychometric Tools. In *Proceedings of the 33rd ACM Conference on User Modeling, Adaptation and Personalization* (pp. 323-328).

Damasio, A. R. (1994). *Descartes' error: Emotion, reason, and the human brain*. G. P. Putnam's Sons.

Dautenhahn, K. (2007). Socially intelligent robots: dimensions of human–robot interaction. *Philosophical transactions of the royal society B: Biological sciences*, *362*(1480), 679-704.




De Waal, F. (2019). *Mama's last hug: Animal emotions and what they tell us about ourselves*. W. W. Norton.

Ekman, P. (1992). An argument for basic emotions. *Cognition and Emotion*, *6*(3–4), 169–200.

Engel, G. L. (1977). The need for a new medical model: A challenge for biomedicine. *Science*, *196*(4286), 129–136.

Gao, L., Jia, Z., Zeng, Y., Sun, W., Zhang, Y., Zhou, W., Zhai, G., & Min, X. (2025). *EEmo-bench: A benchmark for multi-modal large language models on image evoked emotion assessment*. arXiv:2504.16405.

Goleman, D. (1995). *Emotional intelligence: Why it can matter more than IQ*. Bantam Books.

Hu, H., Zhou, Y., You, L., Xu, H., Wang, Q., Lian, Z., Yu, F. R., Ma, F., & Cui, L. (2025). *EmoBench-M: Benchmarking emotional intelligence for multimodal large language models*. arXiv:2502.04424.

Jaggar, A. M. (1989). Love and knowledge: Emotion in feminist epistemology. *Inquiry*, *32*(2), 151–176.

Kambur, E. (2021). Emotional intelligence or artificial intelligence?: Emotional artificial intelligence. *Florya Chronicles of Political Economy*, *7*(2), 147-168.

Lazarus, R. S. (1991). *Emotion and adaptation*. Oxford UP.

Lee, Y. K., Lee, I., Shin, M., Bae, S., & Hahn, S. (2023). *Chain of empathy: Enhancing empathetic response of large language models based on psychotherapy models*. arXiv:2311.04915.

Lindquist, K. A., Wager, T. D., Kober, H., Bliss-Moreau, E., & Barrett, L. F. (2012). The brain basis of emotion: A meta‑analytic review. *Behavioral and Brain Sciences*, *35*(3), 121–43.




Martinez-Plumed, F., & Hernandez-Orallo, J. (2018). Dual indicators to analyze ai benchmarks: Difficulty, discrimination, ability, and generality. *IEEE Transactions on Games*, 12(2), 121-131.

Mayer, J. D., Salovey, P. & Caruso, D. R. (2016). The ability model of emotional intelligence: Principles and updates. *Emotion Review*, *8*(4), 290–300.

Nussbaum, M. C. (2001). *Upheavals of thought: The intelligence of emotions*. Cambridge University Press.

Paech, S. J. (2023). *EQ-bench: An emotional intelligence benchmark for large language models*. arXiv:2312.06281.

Panksepp, J. (1998). *Affective neuroscience: The foundations of human and animal emotions*. Oxford UP.

Parks, M. (2025). The ethics of AI at the intersection of transgender identity and neurodivergence. *Discover Artificial Intelligence*, 5(1), 34.

Pérez-González, J.-C., Perakakis, P., & Petrides, K. V. (2020). Trait emotional intelligence: Foundations, assessment, and education. *Frontiers in Psychology*, *11*, Article 593720.

Plutchik, R. (1980). A general psychoevolutionary theory of emotion. In *Theories of Emotion* (pp. 3-33). Academic Press.

Picard, R. W. (1997). *Affective computing*. MIT Press.

Russell, J. A. (1980). A circumplex model of affect. *Journal of Personality and Social Psychology*, *39*(6), 1161–1178.

Ryle, G. (1949). The Concept of Mind. London: Hutchinson.

Sabour, S., Liu, S., Zhang, Z., Liu, J. M., Zhou, J., Sunaryo, A. S., Lee, T. M. C., Mihalcea, R., & Huang, M. (2024). *EmoBench: Evaluating the emotional intelligence of large language*




*models*. In Proceedings of the 62nd Annual Meeting of the Association for Computational

Linguistics (ACL 2024, Long Papers), 5986–6004.

https://doi.org/10.18653/v1/2024.acl-long.326

Schuhmann, C., Kaczmarczyk, R., Rabby, G., Friedrich, F., Kraus, M., Kalyan, K., Nadi, K.,

Nguyen, H., Kersting, K., & Auer, S. (2025a). *EmoNet-face: An expert-annotated

benchmark for synthetic emotion recognition*. arXiv:2505.20033.

Schuhmann, C., Kaczmarczyk, R., Rabby, G., Friedrich, F., Kraus, M., Nadi, K., Nguyen, H.,

Kersting, K., & Auer, S. (2025b). *EmoNet-voice: A fine-grained, expert-verified

benchmark for speech emotion detection*. arXiv:2506.09827.

Shibata, T. (2012). Therapeutic seal robot as biofeedback medical device: Qualitative and

quantitative evaluations of robot therapy in dementia care. *Proceedings of the IEEE*,

*100*(8), 2527-2538.

Singer, T., & Lamm., C. (2009). The social neuroscience of empathy. *Annals of the New York

Academy of Sciences*, *1156*, 81–96.

Scherer, K. R. (2009). The dynamic architecture of emotion: Evidence for the component process

model. In D. Sander & K. R. (Eds.), *The Oxford companion to emotion and the affective

sciences* (pp. 142–143). Oxford UP.

Zadeh, A., Chan, M., Liang, P. P., Tong, E., & Morency, L.-P. (2019). *Social-IQ: A question

answering benchmark for artificial social intelligence*. In Proceedings of the IEEE/CVF

Conference on Computer Vision and Pattern Recognition (pp. 8807–8817).

Zhang, B., Ma, R., Jiang, Q., Wang, P., Chen, J., Xie, Z., Chen, X., Wang, Y., Ye, F., Li, J., Yang,

Y., Tu, Z., & Li, X. (2025). *Sentient agent as a judge: Evaluating higher-order social

cognition in large language models*. arXiv:2505.02847.